\documentclass[journal]{IEEEtran}
\usepackage{graphicx}
\usepackage{adjustbox}
\usepackage{booktabs, multirow, tabularx, threeparttable}
\usepackage{multirow}

\ifCLASSINFOpdf
\else
\fi
\begin{document}
%
\title{Extending Memory for Language Modelling}
%
%
%

\author{Anupiya Nugaliyadde
\thanks{Anupiya Nugaliyadde is with idoba and Murdoch University, Australia, e-mail: (anupiya.nugaliyadde@idoba.com, a.nugaliyadde@murdoch.edu.au).}
}

%
%

\markboth{Journal of \LaTeX\ Class Files,~Vol.~14, No.~8, August~2015}%
{Shell \MakeLowercase{\textit{et al.}}: Bare Demo of IEEEtran.cls for IEEE Journals}
%



\maketitle

\begin{abstract}
Breakthroughs in deep learning and memory networks have made major advances in natural language understanding. Language is sequential and information carried through the sequence can be captured through memory networks. Learning the sequence is one of the key aspects in learning the language. However, memory networks are not capable of holding infinitely long sequences in their memories and are limited by various constraints such as the vanishing or exploding gradient problem. Therefore, natural language understanding models are affected when presented with long sequential text. We introduce Long Term Memory network (LTM) to learn from infinitely long sequences. LTM gives priority to the current inputs to allow it to have a high impact. Language modeling is an important factor in natural language understanding. LTM was tested in language modeling, which requires long term memory. LTM is tested on Penn Tree bank dataset, Google Billion Word dataset and WikiText-2 dataset. We compare LTM with other language models which require long term memory.
\end{abstract}


%
\IEEEpeerreviewmaketitle

\section{Introduction}
%
%
%
%
\IEEEPARstart{N}{atural} language holds sequential patterns, that connects the past information to the current and future context \cite{mnih2009improving}  \cite{chen2018sequential}. Similar to humans, machine learning models use past sequential context to understand language \cite{shi2015contextual}. A machine learning model usually captures the past sequence in a context to understand the language \cite{huang2012improving}. Holding long sequential context in memory and relating the information in a machine learning model is important to understand context.

Language modelling learns the sequential pattern in natural language to understand and learn the language. The sequential knowledge 
 
Long sequential memory allows machine learning model to relate and extract information in order to understand the context. Deep learning models are capable of holding long sequences and identifying relationships and patterns in a sequence \cite{lecun2015deep}. However, deep learning models are not capable of holding an infinitly long sequences \cite{nugaliyadde2019language}. Therefore, deep learning models have not been as successful in language understanding as in image processing \cite{young2018recent} \cite{worsham2020multi}. A deep learning model which is capable of capturing longer sequences has a potential to improve language understanding. Capturing the past sequence and predicting the next sequence is used to evaluate a machine learning model's capability of understanding natural language using the past sequence \cite{dietterich2002machine}. 

Memory networks have shown to outperform other deep learning models for sequential data \cite{luo2020learning} \cite{khademi2020multimodal} \cite{nugaliyadde2019language}. Recurrent Neural Networks (RNNs) hold the key concept of learning from sequential data. RNNs combine the past inputs with the current inputs to generate the output. Equation 1 demonstrates that the current input $x_t$ is combined with the output of the previous step $y_{t-1}$ to generate the current output $y_t$.  Therefore, $y_t$ not only depends on $x_t$ but also depends on $y_{t-1}$. $W_t$ represents the weights involved in the RNN for a given time frame $t$. The RNN fucntion is represented by $f$. This can be considered as a fundamental approach for holding memory. However, taking $x_{t-1}$ into RNN continously may lead to exploding or vanishing gradient problem. This problem was mainly due to the overlap of the RNN weights causing the  RNN to fail \cite{gers1999learning}.  
\begin{equation}
y_t=f (y_{(t-1)}  ,x_t, W_t)
\end{equation}
Long Short Term Memory (LSTM) network introduced a gating structure to avoid the exploding and vanishing gradient problem \cite{gers1999learning}. This gating structure ensures that the LSTM's weights would not be overloaded. This is controlled by the forget gate's mechanism in the LSTM (2). Equation 2 demonstrates the functionality forget gate. $W_f$ is the weight assigned to the forget gate and adjusts to forget the past sequence. $h_{t-1}$  is the cell state that is passed on from the previous output$y_{t-1}$. $x_t$ is the current input. $h{t-1}$ and  $x_t$ are combined to create the input to the forget gate. $b_f$ is the bias added to the forget gate to give the bias to the LSTM. Based on these parameters the forget gate decides to forget the past sequence or carry forward the past sequence.

 The forget gate decides when to forget the past sequence. The forget gate decides this based on the current input. The forget gate would decide if the current input requires the past outputs, if not the past outputs would be learnt to be forgotten. 

\begin{equation}
f_t= \sigma (W_f.[h_{t-1},x_t] + b_f)
\end{equation}

LSTMs forget gate, removes the unwanted sequential information from the memory. This helps the LSTM’s gradients to avoid vanishing or exploding and has been effective in learning long sequential data. LSTM’s performance for a sequence of 100 steps or more as the input is suboptimal because the forget gate removes entire past sequences \cite{mikolov2014learning}. The forget gate learns to remove the past sequences when the sequence becomes more irrelevant \cite{mikolov2014learning}. However, in language context can carry long dependencies, which can span throughout a very long sequence. Therefore, in long sequences language modelling LSTM's performance can be suboptimal. Variations of the LSTM were introduced however, the method of avoiding vanishing or exploding gradient was not changed \cite{quan2021holistic} \cite{santhanam2020context} \cite{krause2016multiplicative}. The variations of LSTM focused more handling the different input and producing various outputs\cite{mikolov2014learning}. These variations have not improved the suboptimal performance of long term memory in LSTMs \cite{zhao2020rnn} \cite{chung2014empirical}. Gated Recurrent Neural Network (GRU) and Simple Recurrent Neural network (SRN) \cite{mikolov2014learning} use gates to handle vanishing or exploding gradient problem. These gates are used to forget the past sequences to handle the exploding or vanishing gradient problem \cite{gluge2014learning}. Gates in these memory networks prevent the vanishing or exploding gradient problem but sacrifices learning long term dependencies in a sequence. The long sequences in language require sequential long term memory.

Learning from long sequences is important for language modelling. However, remembering the past sequences has many challenges; either affected by the vanishing or exploding gradient problem or forgetting the sequences. The proposed Long Term Memory Network (LTM) is capable of learning from short and long sequences without forgetting sections of the sequence or getting affected by exploding or vanishing gradient problem. LTM takes $x_t$ and combines with $y_{(t-1)}$. It does not forget the sequence and  generalizes the $x_t$ and $y_{(t-1)}$. LTM is evaluated on language modeling to demonstrate long term memory capabilities. LTM also shows that it is capable of learning from short term. LTM has shown better performance over other long term memory models for language modeling. 
The main objectives of this paper are:
\begin{enumerate}
\item Introduce and demonstrates the long term and short term learning capabilities of the LTM for language modelling for character level and sentence level.
\item Demonstrate that LTM is capable of handling long sequence without forgetting the past sequences and being affacted by the vanishing or exploding gradient problem.
\end{enumerate}

\section{Related Work}
Sequential data carries knowledge through the sequence and the previous sequences affect the future sequences. Therefore, learning from the sequence is a key factor. Memory networks are commonly used for sequential learning tasks \cite{nugaliyadde2019language}. RNN is an initial model which uses sequential data to learn \cite{mikolov2010recurrent}. Although RNN suffers from the vanishing or exploding gradient problem the key concept of RNN is used in all the memory networks (3). The input ($x_t$) combines the past output ($y_{t-1}$) and generates the output($y_t$). The weights ($W_i$) are adjusted using the activation function. Equation (3) shows that $y_t$ depends on $y_{t-1}$. This concept is used in all the other memory network. 

\begin{equation}
{y_t} = activation(W_i[y_{t-1};x_t])
\end{equation}
\subsection{ Vanishing or Exploding Gradient Problem}
The continous multiplication of $W$ in Equation (3), can cause the gradient back-propagation to grow or decay exponential. Eigenvalues greater than 1 in $W$ causes exploding gradient. On the other hand, if the eigenvalue is less than 1, it can leads to vanishing gradient \cite{chandar2019towards}. A saturating activation function having eigenvalues less than 1 or equal to one can deteriorate in backpropagation. 

\subsection{Avoiding Vanishing or Exploding Gradient Problem}
Gradient clipping limits exploding gradient but vanishing gradient is hard to prevent \cite{bengio1994learning}. Vanishing or exploding gradient problem was avoided in LSTM by introducing a gating structure \cite{gers1999learning}. The gates control the data flow in a neural network model \cite{hochreiter1997long} and forgets the past sequence when it is irrelavant. However, forgetting a sequence can affect the models predictions negatively because the network tends to forget old sequences which the model determines not relavant \cite{tsironi2017analysis} . Language modeling can be used to evaluate memory networks \cite{sundermeyer2012lstm}. Better language models rely on long term memory because the knowledge in of a language is carried through long sequences. Long term memory carries the knowledge through the sequence and is capable of extracting better knowledge \cite{mcnamara1996good}.

RNN is used in various forms for language modelling \cite{mikolov2010recurrent}, \cite{mikolov2011extensions}. These approaches use RNN by improving the memory of the RNN. However, the modified RNNs have shown to suffer from the vanishing and exploding gradient problem with long sequences. In order to improve on language modeling LSTM was introduced as it has shown to handle vanishing and exploding gradient \cite{sundermeyer2012lstm}. 
Forget gate of the LSTM is used to prevent the LSTM have exploding or vanishing gradients, by removing irrelevant past sequences from the LSTMs’ memory. The forget gate also prevents the LSTM from vanishing and exploding gradients by controlling the internal memory and removing the long and irrelevant sequences. LSTM uses the forget gate to prevent $W \approx 0$ and preventing the vanishing gradient (4). When $\frac{\partial E_T} {\partial W}$ reaches 0 the forget gate would forget $W$ from the past sequence. 
\begin{equation}
\frac {\partial E} {\partial W} =\sum_{t=1}^{k+1} \frac{\partial E_T} {\partial W}
\end{equation} 

LSTM's cell state is created by using additive functions, to prevent it from reaching the vanishing gradient. Therefore, the LSTMs cell state can reach higher value. Language sequences can be long and information in a sequence can be carried throughout paragraphs and chapters. Therefore, a language model should be capable of holding long sequences. Various modifications are applied to LSTM in order to support long term memory for language modelling \cite{sundermeyer2012lstm} \cite{merity2017regularizing} \cite{kurata2017language}. Although these architectural changes showed promise, they were unable to avoid the forget gates’ impact on long sequences. The variations of LSTM's changed the gating structures and connections within the LSTM cells. However, these variations used included the forget gate \cite{greff2016lstm}. Therefore, an approach to handle long term memory, which is not effected by the vanishing or exploding gradient problem is required.

New models are introduced to handle long-term memory and short-term memory. The AntisymmetricRNNs’ connect RNN and the differential equations \cite{chang2019antisymmetricrnn}. AntisymmetricRNN uses the differential equations stability property to capture long-term dependencies. AntisymmetricRNN has been shown to outperform the LSTM on long-term memory tasks and matches performance in short-term memory tasks. AntisymmetricRNN has a simpler and smaller architecture compared to the LSTM. However, the AntisymmetricRNN was tested on image sequence data. h-detach is a stochastic algorithm specified to optimize LSTM to improve on long-term memory tasks \cite{arpit2018h}. h-detach prevents the gradients to flow through the cell states. Therefore, the cell state would not suppress the weights and LSTM would capture long dependencies (long-term memory). This has also been tested with image-related datasets and image captioning datasets to test long-term dependencies. Carta et al., propose a Linear Memory network featuring an encoder-based memorization component built with a linear autoencoder for sequences\cite{carta2021encoding}. The encoder-based network is developed to enhance short-term memory. The network is tested on a White noise dataset and sequential image dataset.

Novel models are introduced to handle long-term memory in language modelling. Non-saturating Recurrent Unit (NRU) has avoided saturating the activation function and saturating gates \cite{chandar2019towards}.  Furthermore, NRU uses a rectified linear unit (ReLU) to support the long term memory, with the novel architecture. NRU avoids the vanishing or exploding gradient problem with long term memory. NRU has also shown to have performance similar to that of the gated memory models for short term memory tasks. However, a simple gated approach has shown promising results compared with the NRU \cite{cheng2020refined}. Non-normal Recurrent Neural Network (nnRNN) uses a Schur decomposition for a connectivity structure and avoids computing Schur decomposition and dividing Schur form into normal and non-normal \cite{kerg2019non}. nnRNN uses orthogonal recurrent connectivity matrices with non-normal terms increases the flexibility of a recurrent network. nnRNN has shown to perform well for long term memory tasks and increased expressivity on tasks requiring online computations to transient dynamics \cite{kerg2019non}. nnRNN has demonstrated that connectivity with gates have shown a higher performance in learning and has advantages over LSTM and GRU. However, these models have not shown to learn long sequential patterns that are carried throughout a given long sequence. Learning from long sequences is an important part in natural language understanding because knowledge and their relationships are carried throughout long sentences, paragraphs, chapters and books. Therefore, it is important to focus on a machine learning approach which is capable of learning from long sequences. 

\section{Long Term Memory Network}
LTM is designed to learn from long sequences (more than 300 time stamps) with minimal number of units combined without forgetting the sequence. LTM cell is capable of learning and generalizing from the past outputs. Furthermore, LTM gives a high precedence for the current input ($x_t$). 

\begin{figure}
  \includegraphics[width=\linewidth]{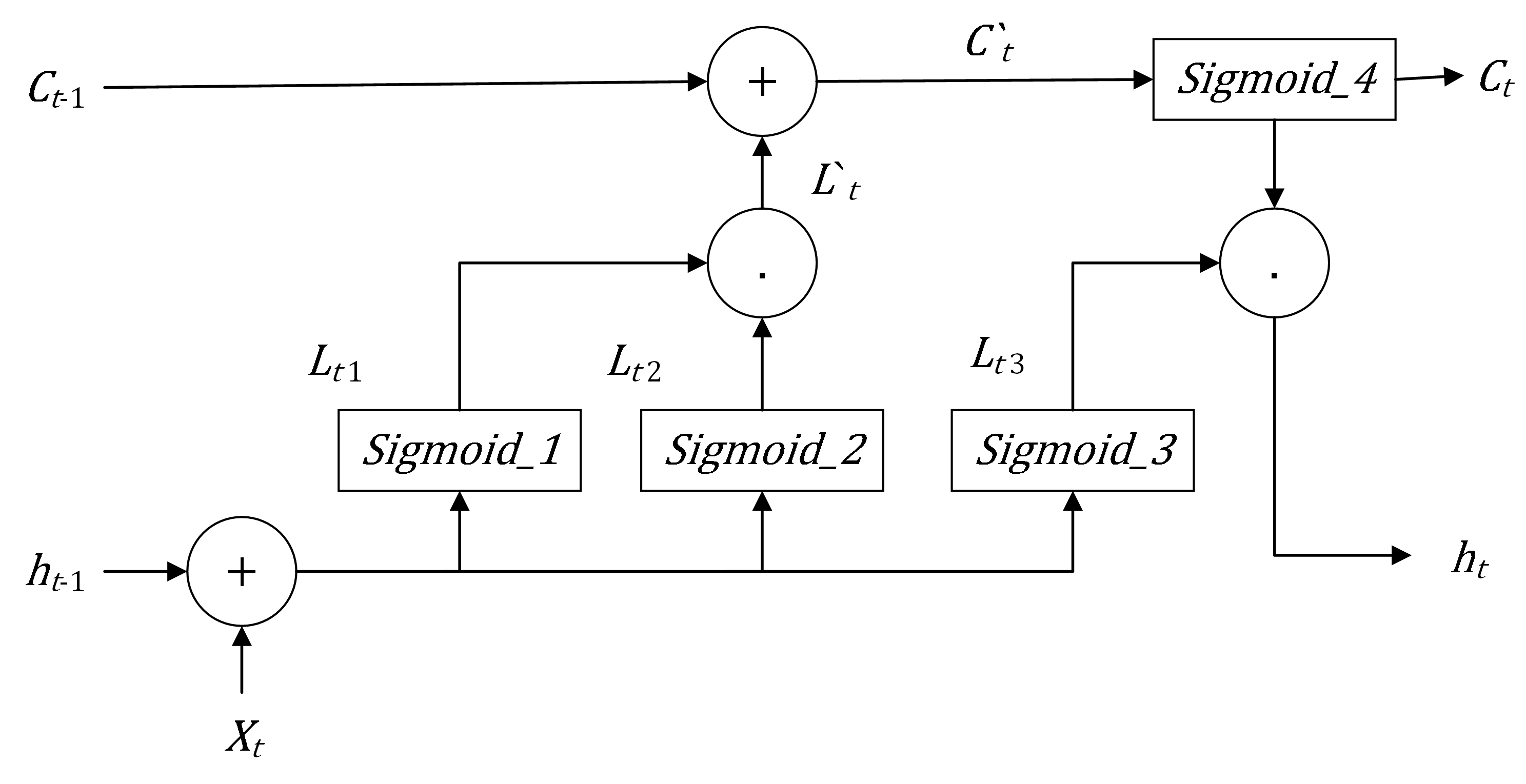}
  \caption{A Long Term Memory Network cell. Data flow is shown by the arrows. }
  \label{fig:LTM}
\end{figure}

LTM cell is divided into three main sections. The sections are:
\begin{enumerate}
\item Input state: this handles the input that is passed into the network
\item Cell state: carries the past processed data and combines them with the currently processed input to carry forward
\item Output state: creates the final output by combining the processed current input with the cell state’s output.
\end{enumerate}
The gates are used in order to give precedence to the current input and prevent the exploding or vanishing gradient. This is different from LSTM's main reasons of using gates for prevent exploding or vanishing gradient \cite{hochreiter1997long}.  The gates play a main role in the structure of the LTM. 
\subsection{Input state}
Input state handles $x_t$ passed to the LTM. Initially, LTM combines the past output ($h_{t-1}$) to $x_t$ as shown in (4). The $\sigma$ indicates the sigmoid functions and $W_1$ is the weight. 
\begin{equation}
L_{t1}= \sigma (W_1(h_{t-1} + x_t))
\end{equation}
$W_2$ is a different weight assigned for $L_{t2}$. $L_{t1}$ and $L_{t2}$ are influenced by the $x_t$. Therefore, $L_{t1}$ and $L_{t2}$ are the direct products of the $x_t$. 
\begin{equation}
L_{t2}= \sigma (W_2(h_{t-1} + x_t))
\end{equation}

Equation (7) combines $L_{t1}$ and $L_{t2}$ in order to create $L_t$. Equation (7) result $L'_{t}$ is used to influence the cell state with an emphasis on $x_t$. $L'_t$ is the dot product of $L_{t1}$ and $L_{t2}$. 
\begin{equation}
 L\\'_{t}= L_{t1}.L_{t2}
\end{equation}

$ L'_t$ amplifies the effects of $x_t$ and the past output $h_{t-1}$. $ L'_t$ is added to the cell state in order to be carried forward for step $t+1$.

\subsection{Cell state}
The cell state is responsible to carry forward from step $t-1$ to step $t$. This carry forward information is required in order to use $x_{t-1}$to support predicting $h_t$. The cell state carries the past inputs in order to hold the sequential information from $t-n$ to $t$. $ L'_t$ is added to the cell state to pass the current input to the next step $t+1$. Equation (7) shows the combination of $C_{t-1}$ to $L'_t$. 
\begin{equation}
 C\\'_{t}= L\\'_t + C_{t-1}
\end{equation}

$C'_t$ would be influenced by $x_t$ and passed on to step $t+1$. $x_t$ influences $ C'_t$ with a higher weight due to (7). However, if the cell state is overloaded with sequential data without control, LTM can be affected by the exploding or vanishing gradient \cite{pascanu2012understanding}. Concept of LTM is based on holding long sequences. Therefore, as shown in (8), the cell state is scaled. This prevents the LTM from reaching an exploding or vanishing gradient by generalizing the output using Sigmoid function. $C_t$ would carry forward the cell state and the past sequential information to the next step.

\begin{equation}
C_t= \sigma (W_4 . C\\'_t)
\end{equation}

\subsection{Output state}
Output state creates the final output ($h_t$) of LTM.  As shown in (9) and (10), $h_t$ is directly influenced by $x_t$. Equation (9) is a key element that decides the final output as shown in (10). 

\begin{equation}
L_{t3}= (W\\'_3 (h_{t-1} + input))
\end{equation}

Equation (10), generates the final output ($h_t$). This is generated by combining both $C_t$ and $L_{t3}$.
\begin{equation}
h_t = C_t . L_{t3}
\end{equation}
 
$C_t$ and $h_t$ are carried on to the next layer of the LTM. Therefore, $C_t$ and $h_t$ are carried forward which influnces on $x_{t+1}$ and generates $h_{t+1}$. 

\subsection{Generalization and Avoiding Vanishing or Explording Gradient in LTM}
LTM presented in \cite{nugaliyadde2019language} is further extended to prove its capability of generalization of the internal cell state value. The term ''generalization'' in this paper is used to discuss the internal values within the LTM cell which is contrary to the normal term of generalization used in deep learning/ machine learning. LTMs' $C_t$ requires the capable of generalizing the past knowledge without forgetting the past sequence (8) and preventing exploding or vanishing gradient in the cell state. Averaging techniques are commonly used to generalize the output. In this paper LTM uses the sigmoid function to generalize $C'_t$ in the cell state which is carried on to $t+1$ as $C_t$. LTM uses the $C'_t$ which is generated using $C_{t-1}$ and the intermediary outcome ($L'_t$) of $x_t$ to generalize the past sequences. The sigmoid function (11) places the input to the function in between 0 to 1. The cell states output (${C_t}$) is the sigmoid functions output, which uses $C'_t$ and $L'_t$. Therefore, the cell state is controlled internally and prevents the cell state expand exponentially. 

\begin{equation}
f_(x) = \frac{1} {1+e^{-x}}
\end{equation}

The sigmoid function on the cell state can be expanded as shown in (12). $C_t$ carries forward the past sequential information that was acquired and combines with the current input information and maintains it between 0 and 1. Therefore, $C_t$ is distributed which generalize the cell states output. This also supports the LTM cells from the vanishing and exploding gradient problem because the internal state is controlled within a controlled area.

\begin{equation}
C_t = \frac{1} {W_4 (L'_t + C_{t-1})}
\end{equation} 

Most memory networks use a gating structure or gradient clipping to prevent the exploding and vanishing gradient problem, which can negatively affect the long-term result in a memory network \cite{chung2014empirical}. LTM manages to keep the past sequences without adversely affecting long term memory of the LTM. Vanishing or exploding gradient occurs in iterative models (13) where $f$ is the iterative function, $x$ is the input and $h^t$ is an activation. When $f$ is iterative the affect can increase exponentially. Therefore, if the output of $f$ becomes $\approx$ 0 or $>$ 1 vanishing or exploding gradient can occur.

\begin{equation}
h^t = f (f(f(h^1,x^1),x^2),x^3)
\end{equation} 
In order to prevent the vanishing/exploding gradient in the LTM, the backpropagated error ($E_t$) for the time ($t$) and the weights ($W$) is as shown in (14) should not be $\approx$ 0 or $>$ 1. 
\begin{equation}
\frac {\partial E} {\partial W} =\sum_{t=1}^{T} \frac{\partial E_T} {\partial W}
\end{equation} 
Therefore, LTM should prevent $\frac{\partial E_T} {\partial W}$ from reaching 0 or $>$ 1. The use of the sigmoid function in creating the $C_t$, prevents the $C_t$ prevents the $C_t$ from $\approx$ 0. Even if the $T$ is very large the sigmoid function would keep the $C_t$ without reaching 0 or above 1. The recursive effect of (13) would be avoided through the sigmoid function. $C_t$ would be kept between the given range (0 and 1), and prevent the values that are passed to $t+1$, reaching 0 or going above 1. Furthermore, the additive property in the cell state prevents the cell state from $\approx$ 0 (7). $C'_t$ this would prevent $C_t$ from $\approx$ 0. The addition increases the cell state value and through the sigmoid function the cell state is kept in between the exponentially increasing or decreasing preventing the vanishing or exploding gradient. 

\section{Model Architecture Comparison to Other Language Models}
LTM has a different architectural design to support long term memory. The structure and the connections in the LTM are different from the other language modeling deep learning models. Therefore, to distinguish the LTM architectural design it is compared with RNN, LSTM, GRU, NRU and the basic Transformers.
\subsection{RNN}
RNN has an architecture which uses the past outputs to generate the current output (1). However, in the RNN there is no additional processing. RNN continously passes the previous output with the current output. However, vanishing and exploding gradient problem affects the RNN.

\subsection{Vanilla LSTM}
LSTM is one of the most commonly used memory networks. LSTM introduced a gating structure to avoid exploding or vanishing gradient problem \cite{gers1999learning}. The gates structure are used to control the data flow in the LSTM cell. Fig.2. depicts the LSTM structure. The forget gate (2) makes decision of forgetting or remembering the past data. However, LSTM does not carry long term memory because it forgets the past data, depending on the current input.
\begin{figure}
  \includegraphics[width=\linewidth]{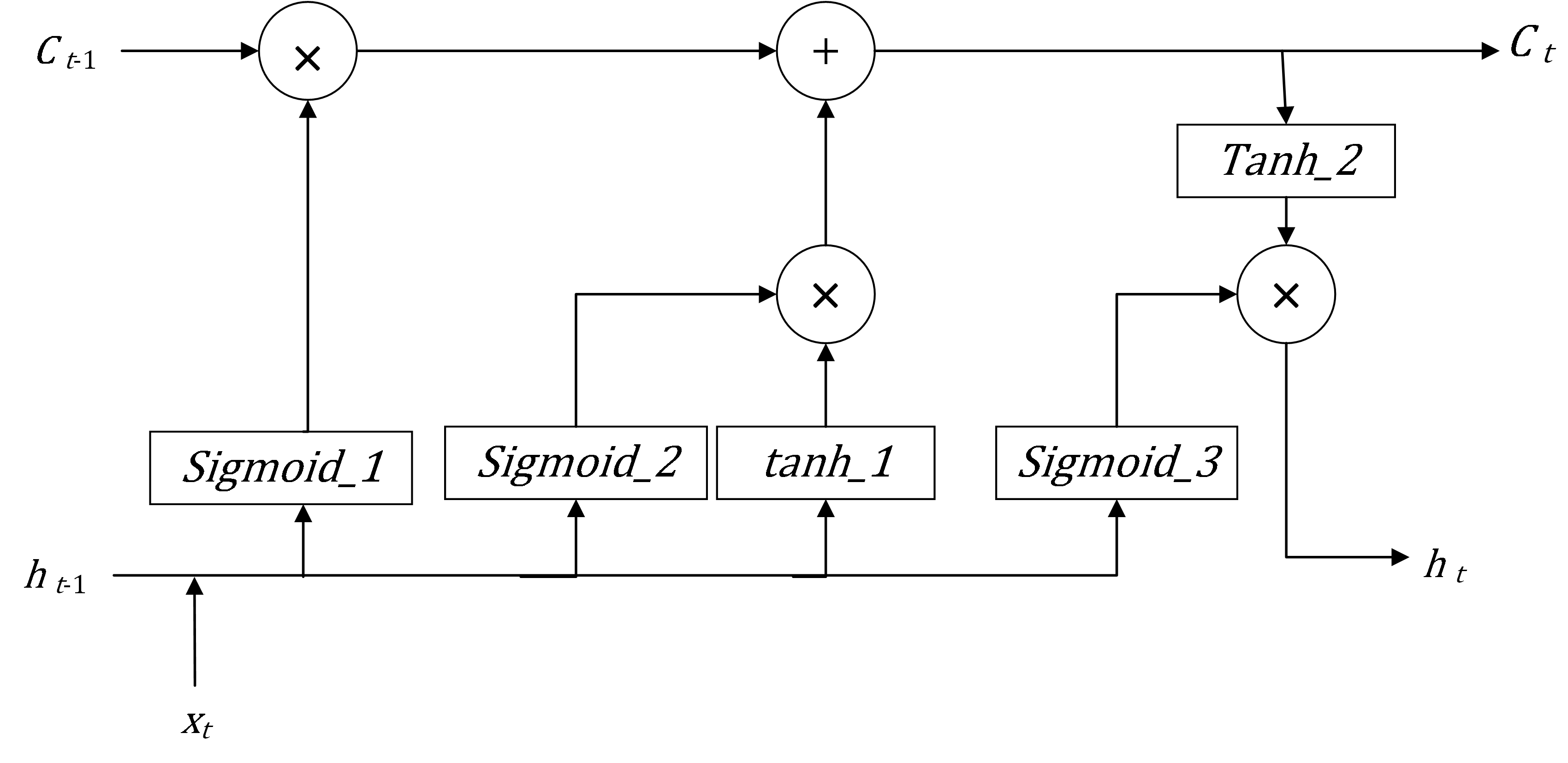}
  \caption{A Long Short Term Memory Network cell. Data flow is shown by the arrows. }
  \label{fig:LSTM}
\end{figure} 

\subsubsection{Comparison Between LSTM and LTM}
Long Short-Term Memory Network (LSTM) (Fig.2.) and Long-Term Memory Network (LTM) (Fig.1.) have similarities and differences. The Table \ref{table:20} compares LTM and LSTM. 

\begin{table*}[h]
\centering
\begin{tabular}{@{}p{3cm}p{3cm} p{3cm}lll@{}}
\hline
Components of the Memory Networks & LTM & LSTM \\ [0.5ex] 
 \hline
 Forget gate  & Does not have a forget gate. Therefore, does not remove any past sequences.  & Has a forget gate to remove past sequences which are not relevant to the current input. 
The forget gate prevents the LSTM from overloading the memory. \\ \hline
Input and Output gates & Input and output gates handles the inputs and outputs to the LTM. & Input and output gates handles the inputs and outputs to the LSTM. \\ \hline
 Cell state  & Cell state carries the past sequence forward to be added to next input.
However, the current input is given a higher precedence when before passing to the cell state.  & Cell state carries the past sequence forward to be added to the next input.     \\ \hline
Activation functions & Only uses sigmoid activation function. The sigmoid function scales the sequence.  & There are combinations of sigmoid and tanh function.  \\
 \hline
\end{tabular}
\caption{Comparison between LTM and LSTM\'s Components.}
\label{table:20}
\end{table*}

\subsection{Variations of LSTM}
LSTM is adjusted to fit for various different tasks \cite{zhao2018sequential}. However, these adjustments did not change the core-structure of the LSTM as shown in Fig.3, which is one modification that has a peephole connection to the gates. This allows the network to learn the fine distinction between spikes. This directly supports long term memory. The main contribution of this network was to use the time counter which enables the network to count the number of steps taken and predict at the required time step. However, the network only remembers 50$^{th}$ steps and not the data in-between. It does not hold all information. Therefore, long term dependencies which require the entire sequence has a higher probability to fail. 

\begin{figure}
  \includegraphics[width=\linewidth]{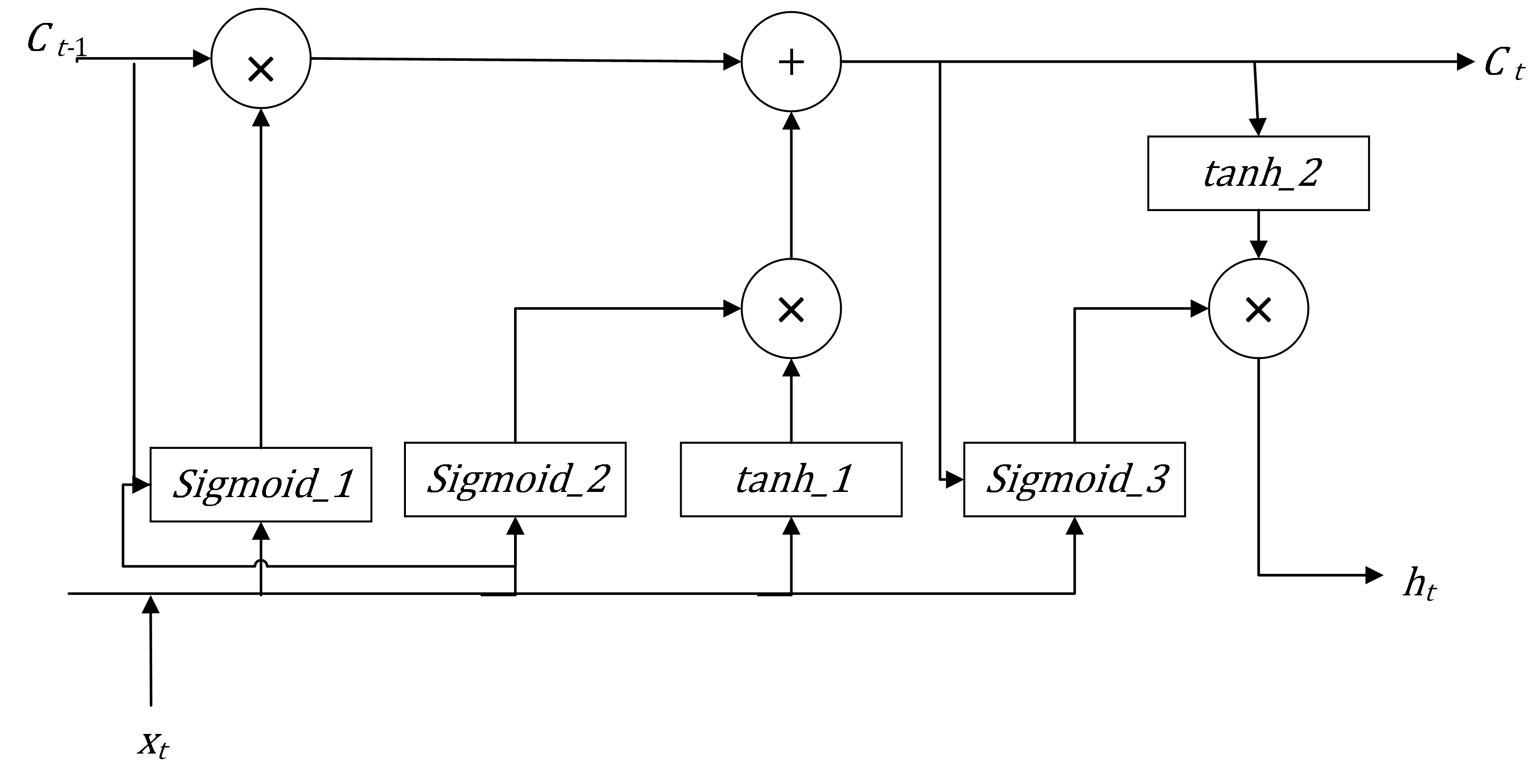}
  \caption{A variant of Long Short Term Memory Network cell. Data flow is shown by the arrows. }
  \label{fig:LSTMV}
\end{figure} 

\subsection{GRU}
GRU has a simple architecture, which has simple set of gates (Fig.4.). GRU was introduced to prevent the vanishing and exploding gradient problem. GRU has a forget gate and a reset gate which decides if the input will be transferred to the output. 

\begin{figure}
  \includegraphics[scale=0.5]{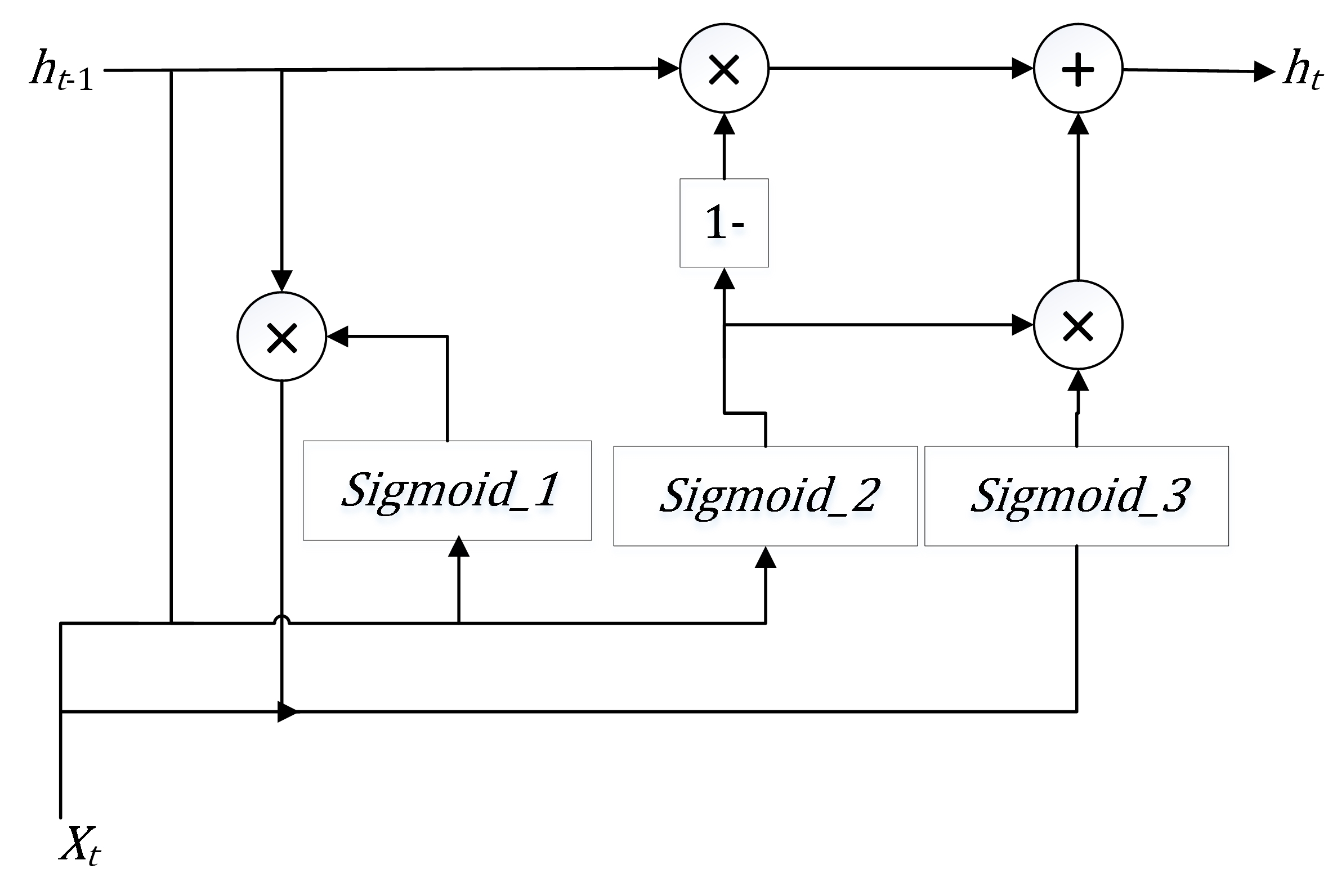}
  \caption{A Gated Recurrent Unit. The data flow is shown in the arrow. }
  \label{fig:GRU}
\end{figure} 

\subsection{Transformers}
The core of transformers is based on attention and uses an encoder and decoder architecture \cite{vaswani2017attention}. The transformer comprises of a stack of encoders that take the input and a stack of decoders that produce the output. The transformer-based language models are decoder stacks in transformers. Generative Pre-trained Transformer (GPT-3) is one of the early and popular transformers which uses 175 billion parameters for language modeling \cite{brown2020language}. GPT-3 is a set of decoders comprised of self-attention layers and a feed-forward neural network for the final output. Therefore, the transformers are highly reliant on the attention throughout the sequence of the context. However, the unidirectional transformers are not capable of capturing both left and right context in all the layers. This adversely affects the transformer's capabilities.

Bidirectional Encoder Representations from Transformers (BERT), is a bidirectional transformer that is used for natural language understanding tasks \cite{devlin2018bert}. BERT trains and gives a deep sense of language context and flow through its bidirectional nature compared to the other unidirectional transformers. However, BERT uses the attention mechanism that learns contextual relations between words in a context and creates Masked Language Modelling which allows bidirectional training. This capability allows BERT to understand the context using both the left and the right context to a given word and generate a better language model. BERT is a large neural network with 24 transformer blocks, 1024 hidden layers, 16 self-attention heads, and uses 340 Million parameters. Therefore, BERT (and most of the transformers for language modeling) utilize the GPUs to pass the inputs parallel to each other that increasing the training speed of the transformers. Therefore, the transformers heavily rely on the use of GPUs and large memory. Training BERT or any transformer-based large language models is challenging. Although a pre-trained BERT model is available for general language modeling, re-training and training large transformer-based language models for specific language models are not practically achievable.

\section{Experiments and Results}
LTM was tested on language modeling tasks which require long term memory. LTM is tested on PennTree Bank (PTB) dataset, Google Billion Words, and WikiText-2 for word wise language modeling. It is also tested on a character level language modeling for PennTree bank dataset. LTM was compared against a number of popular memory networks including various RNN models. 

\subsection{Datasets}
\subsubsection{PennTree Bank Dataset}
PTB contains a wide range of text including text from Wall Street Journal, nursing notes, IBM computer manuals, transcribed telephone conversations and etc. PTB has a 10K vocabulary. This dataset is one of the most popular datasets for language modeling. PTB consists of 930K tokens for training, 74K tokens for validation and 82K tokens for testing \cite{taylor2003penn}. 
\subsubsection{Google Billion Words Dataset}
This dataset consists of 0.8 Billion words for training and testing \cite{chelba2013one}. The data is taken from the WMT11 website. The duplicate sentences are removed, therefore unique sentences are available in the dastaset. The vocabulary is 793471.
\subsubsection{WikiText-2 Dataset}
WikiText-2 holds the complete text (without any filtering applied), with punctuation and numbers \cite{merity2016pointer}. The dataset is composed of complete articles. Therefore, it is suited for long term dependencies.  

\subsection{Character Level Language Modeling}
Predicting the next character(s) in a word sequence of word is character level language modeling. Learning the letter sequence of a word is learnt through short term memory. Therefore, character level language models use short term memory. Table \ref{table:1} empirically demonstrates that LTM is capable of achieving short term memory. 
Penn Treebank Corpus (PTB) is used for character level language modeling. LTM takes the first character in a sequence and predicts the next character. In order to fairly compare LTM to the other memory network models the following setup is used.
The batch size is set to 128. The models were trained for 20 epochs. Character level language modeling is tested on the testing dataset and evaluated with the Bits Per Character (BPC) and accuracy. A lower BPC model is better for predicting the next character. Although character level language modeling does not require long term memory, LTM is tested to compare it with general memory tasks. Table \ref{table:1} shows the test results for character level language modeling for PTB. LTM has not shown higher improvement on the results on character level language modeling but has achieved similar results to standard memory networks such as GRU and LSTM. Table \ref{table:1} shows that LTM generates comparable results with much popular memory network models. 

\begin{table}[h!]
\centering
\begin{tabular}{c c c c} 
 \hline
 Model & BPC & Accuracy \\ 
 \hline
 LSTM & 1.48 & 67.98 \\ 
RNN & 1.55 & 68.43 \\
 GRU & 1.45 & 69.07    \\
JANET & 1.48 & 68.5 \\
NRU \cite{chandar2019towards}& 1.47 & 68.48 \\ 
nnRNN \cite{kerg2019non}& 1.49& - \\
\textbf{LTM} & \textbf{1.44} & \textbf{68.01} \\
 \hline
\end{tabular}
\caption{Testing BPC and Accuracy for Character Level Language modeling for PTB. Results in bold are the LTM's results.}
\label{table:1}
\end{table}

\subsection{Language Modeling}
Language modeling requires a long term memory, especially for sequences which are longer than 50, words. LTM is tested on PTB, Google Billion Words and WikiText-2. The models were evaluated using perplexity. PTB is used as benchmark dataset. Many approaches are tested on PTB. Furthermore, various optimization approaches are tested on LSTM for PTB and WikiText-2 \cite{merity2017regularizing}. In order to fairly compare LTM to other memory network approaches, the same evaluation and testing methods were followed as \cite{merity2017regularizing}. All LTM have 3 LTM layers and 1150 hidden units. Embedding has a size of 400. The batch size for PTB size is 40 and WikiText-2 is 80. The comparison results with LTM on PTB is shown in Table \ref{table:2} .  Furthermore, LTM is tested on WikiText-2 on language modeling (Table \ref{table:3} ).  In order to evaluate long sequential learning with large datasets, LTM was tested on the Google Billion Word test set (Table \ref{table:4}). 

\begin{table}[h!]
\centering
\begin{tabular}{p{5cm} | c c} 
\hline
 Model & Validation & Test \\ [0.5ex] 
 \hline
RNN-LDA + \\ KN-5 + cache \cite{mikolov2012context} & - & 92.0 \\ 
LSTM (large) \cite{zaremba2014recurrent}  & 82.2 & 78.4 \\
Variational LSTM (large, MC) \cite{gal2016theoretically} & - & 73.4    \\
CharCNN \cite{kim2016character} & - & 78.9 \\
Variational LSTM (tied) \\ + augmented loss \cite{inan2016tying} & 71.1 &68.5 \\
Variational RHN (tied) \cite{zilly2017recurrent} & 67.9 & 65.4 \\
NAS Cell (tied) \cite{zoph2016neural} & - &62.4 \\
4-layer skip connection LSTM (tied) \cite{melis2017state} & 60.9 & 58.3 \\ 
AWD-LSTM - 3-layer LSTM (tied) \\ + continuous cache pointer \cite{merity2017regularizing} & 53.9 & 52.8 \\ 
LSTM+ Dual Channel Class Hierarchy \cite{shi2020dual} &- &118.3 \\
LSTM(Large) + cell \cite{qin2020improving} & 76.15 & 73.87 \\
AWD-FWM \cite{schlag2020learning1} &56.76&54.48 \\
\textbf{LTM} & \textbf{52.1} & \textbf{51.7} \\
 \hline
\end{tabular}
\caption{Testing Perplexity for Language modeling on PTB. The best results on each model is reported and LTM results are averaged from 10 different runs.}
\label{table:2}
\end{table}

\begin{table}[h!]
\centering

\begin{tabular}{p{5cm} | c c} 
 \hline

 Model & Validation & Test \\ [0.5ex] 
 \hline
Variational LSTM (tied) + augmented loss \cite{inan2016tying}& 91.5 &87.0 \\
LSTM + continuous cache pointer \cite{grave2016improving} & - & 68.9 \\
NAS Cell (tied) \cite{zoph2016neural} & - &62.4 \\
2-layer skip connection LSTM (tied) \cite{melis2017state} & 68.6 & 65.9 \\ 
AWD-LSTM - 3-layer LSTM (tied) + continuous cache pointer \cite{merity2017regularizing}& 53.8 & 52.0 \\ 
LSTM(Large) + cell \cite{qin2020improving} & 90.52 & 85.76 \\
AWD-FWM \cite{schlag2020learning1} &63.98& 61.65 \\
\textbf{LTM} & \textbf{51.5} & \textbf{50.1} \\
 \hline
\end{tabular}
\caption{Testing Perplexity for Language modeling on WikiText-2. The best results on each model is reported and LTM results are averaged from 10 different runs.}
\label{table:3}
\end{table}

\begin{table}[h!]
\centering
\begin{tabular}{c | c } 
 \hline
 Model & Test Perplexity\\ [0.5ex] 
 \hline
Sigmoid-RNN-2048 \cite{ji2015blackout}& 68.3  \\
Interpolated KN 5-Gram \cite{chelba2013one} & 67.6 \\
Sparse Non-Negative Matrix LM \cite{shazeer2014skip} & 52.9 \\
LSTM-2048-512  \cite{jozefowicz2016exploring} & 43.7  \\ 
LSTM-2048 \cite{grave2017efficient} & 43.9\\
2-layer LSTM-2048 \cite{grave2017efficient} & 39.8\\
GCNN-13 \cite{dauphin2017language} & 38.1 \\
GCNN-14 Bottleneck \cite{dauphin2017language} & 31.9\\
BIG LSTM+CNN inputs \cite{jozefowicz2016exploring} & 30.0\\
BIG GLSTM-G4 \cite{kuchaiev2017factorization}& 23.3 \\ 
\textbf{LTM} & \textbf{21.5} \\
 \hline
\end{tabular}
\caption{Results on the Google Billion Word test perplexity. The best results on each model is reported and LTM results are averaged from 10 different runs.}
\label{table:4}
\end{table}

\subsection{Transformers}
Transformers have outperformed most of the Natural Language Processing tasks. However, language modelling requires memory and sequential information. Transformers are computationally efficient compared to memory networks. However, the transformers are suboptimal for language modelling \cite{wang2019language} \cite{dowdell2020language} . Self-attention and positional encoding in transformers have not incorporated the word level sequential context. LTM is evaluated against the transformers in Table \ref{table:11} shows that LTM outperforms BERT and GTP-3 in language modelling. 

\begin{table}[]
\begin{tabular}{|l|llllll|}
\hline
       &                       &                       &      Datasets                 &                       &                       &  \\ \cline{2-7} 
Models &          PTB             & \multicolumn{1}{l|}{} &           WT-2            & \multicolumn{1}{l|}{} &          WT-103             &  \\ \hline
 & \multicolumn{1}{l|}{Val} & \multicolumn{1}{l|}{Test} & \multicolumn{1}{l|}{Val} & \multicolumn{1}{l|}{Test} & \multicolumn{1}{l|}{Val} & {Test} \\ \hline
GPT-3  & \multicolumn{1}{l|}{79.44} & \multicolumn{1}{l|}{68.79} & \multicolumn{1}{l|}{89.96} & \multicolumn{1}{l|}{80.6} & \multicolumn{1}{l|}{63.07} &{63.47}  \\ \hline
BERT   & \multicolumn{1}{l|}{72.99} & \multicolumn{1}{l|}{62.4} & \multicolumn{1}{l|}{79.76} & \multicolumn{1}{l|}{69.32} & \multicolumn{1}{l|}{109.54} &{107.3}  \\ \hline
LTM    & \multicolumn{1}{l|}{52.1} & \multicolumn{1}{l|}{51.7} & \multicolumn{1}{l|}{51.5} & \multicolumn{1}{l|}{50.1} & \multicolumn{1}{l|}{49.3} & {47.1} \\ \hline
\end{tabular}
\caption{Perplexity comparison on BERT and GTP against LTM for language modelling.}
\label{table:11}
\end{table}

\subsection{Combining Transformers with Memory Networks}
Transformers underperform in Language modeling. Therefore, memory networks are added to the Transformers to better capture sequential knowledge \cite{wang2019language}.  Adding LSTM layers to transformers capture the sequential context and perform efficiently \cite{wang2019language}. Therefore, BERT or GPT-3 coupled with LSTM layers performs well in language modeling.  \cite{wang2019language} uses the LSTM to capture the sequential information and added a Coordinate Architecture Search to find an effective architecture through iterative refinement of the model. The LTM was combined with the Transformers (BERT and GPT). This approach is applied to compare the LTM with novel transformer models. However, the paper's main focus is on the LTM without any other model combination.  Table \ref{table:21} compares the LTM combined with transformers to Coordinate Architecture Search (CAS) with LSTM and transformers.

\begin{table}[]
\begin{tabular}{|l|llllll|}
\hline
       &                       &                       &      Datasets                 &                       &                       &  \\ \cline{2-7} 
Models &          PTB             & \multicolumn{1}{l|}{} &           WT-2            & \multicolumn{1}{l|}{} &          WT-103             &  \\ \hline
 & \multicolumn{1}{l|}{Val} & \multicolumn{1}{l|}{Test} & \multicolumn{1}{l|}{Val} & \multicolumn{1}{l|}{Test} & \multicolumn{1}{l|}{Val} & {Test} \\ \hline
BERT-CAS  & \multicolumn{1}{l|}{39.97} & \multicolumn{1}{l|}{34.47} & \multicolumn{1}{l|}{38.43} & \multicolumn{1}{l|}{34.64} & \multicolumn{1}{l|}{40.70} &{39.85}  \\ \hline
GPT-CAS   & \multicolumn{1}{l|}{46.24} & \multicolumn{1}{l|}{40.87} & \multicolumn{1}{l|}{50.41} & \multicolumn{1}{l|}{46.62} & \multicolumn{1}{l|}{35.75} &{34.24}  \\ \hline
BERT-Large-CAS    & \multicolumn{1}{l|}{36.14} & \multicolumn{1}{l|}{31.34} & \multicolumn{1}{l|}{37.79} & \multicolumn{1}{l|}{34.11} & \multicolumn{1}{l|}{19.67} & {20.42} \\ \hline
BERT-LTM   & \multicolumn{1}{l|}{32.2} & \multicolumn{1}{l|}{30.11} & \multicolumn{1}{l|}{34.8} & \multicolumn{1}{l|}{30.61} & \multicolumn{1}{l|}{16.31} &{14.2}  \\ \hline
GPT-LTM    & \multicolumn{1}{l|}{41.32} & \multicolumn{1}{l|}{37.17} & \multicolumn{1}{l|}{43.1} & \multicolumn{1}{l|}{39.66} & \multicolumn{1}{l|}{34.2} & {33.2} \\ \hline

\end{tabular}
\caption{Combining Transformers to the LTM for Language modelling compared with the Transformer based language models. }
\label{table:21}
\end{table}

\section{Discussion}
Information is sequential and more information is gained with longer sequences. Longer the sequential information that the model is capable of holding better the predictions based on the sequence. However, in long sequences, the immediate prior sequences should have a higher precedence over the older sequence, because the relationship that the prior sequence is more related to the current input than the older sequence. Natural language sequences are influenced by the past sequences but the prior sequence is more related to the current input. Therefore, LTM focuses on giving a high precedence to the prior sequence to predict on the current input compared to the older sequences. The older sequences have a minimal reference to the current input while the prior sequences have a higher precedence on the current input. This factor is utilized in RNN and LSTM to predict the short-term sequences in language modelling. LSTM\'s capabilities of language modelling is better because the predictions are based on prior sequence. The LTM is capable of learning from sequences which are longer than 250 words and has achieved the perplexity scores in Table I, III, IV and V. LTM has achieved the better perplexity compared to the other models because LTM is capable of holding long sequences in its memory.

LTM is tested on multiple language modelling dataset. Language modeling requires memory to learn long term  dependency to predict the next word in a sequence \cite{kuchaiev2017factorization}. Learning dependencies are required to predict the next words. LTM learns the sequence of a given text. Unlike other memory models, LTM is structured to learn long sequences, which can exceed 250 words using less than 10 LTM units. In learning a sequence, LTM is not affected by the vanishing or exploding gradient problem. LTM has a cell state which carries out the past sequential information and carries forward with the hidden state \cite{pascanu2013difficulty}. The use of the cell state is commonly used in memory networks to carry forward the past sequential information \cite{hochreiter1997long}. Furthermore, LTM does not forget the past sequences at any point in the sequence, similar to other common networks such as LSTM and GRU. LTM uses the gate structure to generalize the sequence and give higher priority to the current input and not to control and forget the sequence. LTM continues to learn through the long sequence. As shown in Fig. 1, $Sigmoid\_1$ and $Sigmoid\_2$ are used to give a high priority to the current input. Furthermore, the $Sigmoid\_4$ generalizes and combines the cell state to carry forward the past outputs. Therefore, as shown in Fig. 5, LTM can be set to continue to run for many iterations after convergence and the testing perplexity does not change. LTM is trained for a range of 100 epochs to 5000 epochs and tested to show that LTM has saturated in learning and does not change with the increasing epoches. This shows that the LTM does not get effected by vanishing or exploding gradient even after training has saturated. 
 
\begin{figure}
  \includegraphics[width=\linewidth]{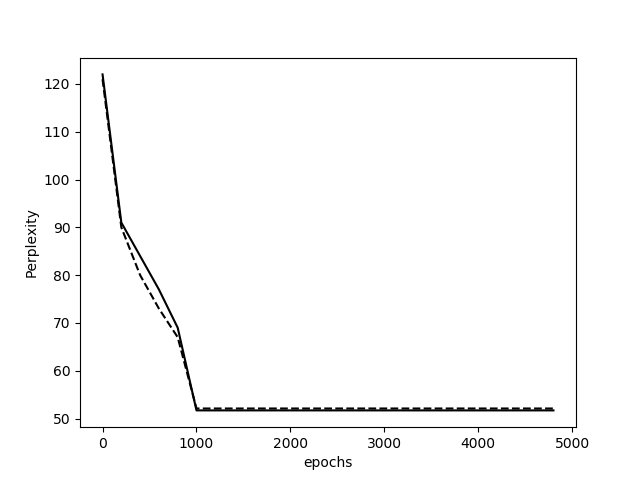}
  \caption{- - shows the training Perplexity, --- shows the testing Perplexity. Training and testing Perplexity change with the number of epochs.}
  \label{fig:graph}
\end{figure}

\subsection{Analysis of the LTM's gating effect}
The LTM's gate alignment and the selection of sigmoid functions is used to support long term learning and giving presidence to the current input. The each gate in the LTM is place to support long term memory. This is demonstrated by removing each gate (setting the gate to generate 1 and pass any input) to and testing the LTM's language modeling capabilities. PTB dataset is used since it is the most common dataset for language modeling \cite{chandar2019towards}. 10 LTM units are used to generate the highest results with the least number of LTM units. The LTM units are combined to demonstrate the impact on each gate towards long term memory in language modeling. Table \ref{table:5}, shows the effect of removing each gate towards the perplexity of predicting a long sequence of 100. It's clear that $Sigmoid\_4$ has the highest impact on the perplexity. $Sigmoid\_4$ directly influences the cell state and the output. Although the $Sigmoid\_1$ has the lowest impact on the perplexity, compared to the LTM the effect of $Sigmoid\_1$ is immense. Table \ref{table:5} clearly shows that the combination of all the gates is required for LTM to produce long term learning. Furthermore, gates are combined togather to show the effect of the set of gates to demonstrate which gates affect long term memory and give precedence on the current input.

 \begin{table}[h!]
\centering
\begin{tabular}{c | c} 
 \hline

 Opened Gate & Test \\ [0.5ex] 
 \hline
Sigmoid\_1 & 90.3 \\
Sigmoid\_2 & 92.2\\
Sigmoid\_3 &99.1 \\
Sigmoid\_4 & 101.1 \\ 
Sigmoid\_4 + Sigmoid\_3 & 111.5 \\
Sigmoid\_1 + Sigmoid\_2 & 120.8 \\
Sigmoid\_1 + Sigmoid\_3 & 87.3 \\
Sigmoid\_1 + Sigmoid\_4 & 81.5 \\
Sigmoid\_2 + Sigmoid\_ 3 & 78.2 \\
Sigmoid\_2 + Sigmoid\_4 & 88.7 \\
Sigmoid\_4 + Sigmoid\_3 + Sigmoid\_2 & 173.9 \\
Sigmoid\_4 + Sigmoid\_3 +  Sigmoid\_1 & 176.2 \\
\textbf{LTM with all gates} & \textbf{51.1} \\
 \hline
\end{tabular}
\caption{Testing Perplexity for Language modeling on PTB for opening each gate. The best results on each model is reported and LTM results are averaged from 10 different runs.}
\label{table:5}
\end{table}

 $Sigmoid\_1$ and $Sigmoid\_2$ gates effects on giving precedence to the current input. Table \ref{table:6} presents the BPC on PTB when both $Sigmoid\_1$ gate and $Sigmoid\_2$ gate are opened (the gate is set to 1) and closed while $Sigmoid\_3$ gate and $Sigmoid\_4$ gates are open. The open gates have no effect on the output. In order to learn short term dependencies character level language modeling is analysed. Table 6 shows that BPC is very high compared with LTM's results, and comparing with Table \ref{table:1} results, $Sigmoid\_1$ and $Sigmoid\_2$ gates have given an upper hand over the other memory networks. Table \ref{table:6} is a clear indication that $Sigmoid\_1$ and $Sigmoid\_2$ gates are responsible for learning short term dependencies. 

 The effects on long term memory is tested using $Sigmoid\_3$ and $Sigmoid\_4$ gates. $Sigmoid\_3$ and $Sigmoid\_4$ gates are opened and LTM is tested on a perplexity of PTB dataset similar to the experiment in Table \ref{table:5}. As shown in Table \ref{table:5} $Sigmoid\_4$ has a high impact on the long term memory. Therefore, combining $Sigmoid\_3$ and $Sigmoid\_4$ gates generates a very high impact on long term memory as shown in Table 7. However, the impact on the current input is high on the predictions, it has affected the perplexity as shown in Table \ref{table:7}. 

\begin{table}[h!]
\centering
\begin{tabular}{c c c c} 
 \hline
 Model & BPC \\ [0.5ex] 
 \hline
 LTM without $Sigmoid\_1$ and $Sigmoid\_2$ gates & 1.71  \\ 
LTM without $Sigmoid\_3$ and $Sigmoid\_4$ gates & 1.49  \\
\textbf{LTM} & \textbf{1.44} \\
 \hline
\end{tabular}
\caption{Testing BPC on LTM to analyse $Sigmoid\_1$ and $Sigmoid\_2$ gates effect on learning short term dependencies.The results are averaged from 10 different runs.}
\label{table:6}
\end{table}

 \begin{table}[h!]
\centering
\begin{tabular}{c | c} 
 \hline

 Gate & Test \\ [0.5ex] 
 \hline
LTM without $Sigmoid\_3$ and $Sigmoid\_4$ gates & 110.3 \\
LTM without $Sigmoid\_1$ and $Sigmoid\_2$ gates & 77.2\\
\textbf{LTM with all gates} & \textbf{51.1} \\
 \hline
\end{tabular}
\caption{Testing Perplexity for Language modeling on PTB for opening $Sigmoid\_3$ and $Sigmoid\_4$ gates to analyse the long term dependency. The results are averaged from 10 different runs.}
\label{table:7}
\end{table}

\subsection{Vanishing and exploding gradient with LTM}
LTM is capable of handling vanishing and exploding gradients. During the training process, the weights of LTM does not reaching infinity or 0. LTM’s weight have not been effected by vanishing or exploding gradient as shown in the Figure 5. Although the LTM receives data continuously the perplexity does not increase or change. If the vanishing or exploding gradients occure the perplexity of the LTM would show a drastic increase because the models prediction would be affected. LTM has been tested even after achieving it’s peak performance and trained and allowed the model to over train, however, even the over trained LTM has not shown that it is affected by the exploding and vanishing gradient problem.

\subsection {LTM's Performance}
LTM is a low-memory CPU and Memory usage model. Table I empirically shows that the LTM can perform well with the low specification with less computational power. The model was tested on a CPU with 8 GB RAM on the PTB dataset.

\begin{table}[h!]
\centering
\begin{tabular}{c | c c} 
\hline
 Sequence Length & Train Time (minutes) & Test Time (seconds) \\ [0.5ex] 
 \hline
50 words &14:32 & $<$0.001 \\ 
100 words  & 14:50 & $<$0.001 \\
200 words & 15:04 & $<$0.001 \\
250 words & 15:17 & $<$0.001 \\
300 words & 15:29 & $<$0.001 \\
350 words & 15:33 & $<$0.001 \\
400 words & 15:40 & $<$0.001 \\ 
450 words & 15:48 & $<$0.001 \\ 
500 words & 15:50 & $<$0.001 \\
600 words & 15:59 & $<$0.001 \\
1000 words &16:10 & $<$0.001 \\
 \hline
\end{tabular}
\caption{Comparison of the sequence length of a LTM on the PTB dataset on an 8 GB RAM CPU computer.}
\label{table:24}
\end{table}

Table \ref{table:24} shows that the LTM learns fast without using a GPU and the learning time does not exponentially grow with the sequence growth. Furthermore, LTM does not require a GPU for training. However, LTM can utilize the GPU to reduce the training time. LTM was trained on an NVIDIA 1080i graphics card and the training time was reduced on average by 7 minutes shown in Table \ref{table:25}. Furthermore, LTM has been tested on a 4 GB RAM computer and the training time increased on average by 4 minutes. Therefore, LTM has not been affected by limited computational resources.

\begin{table}[h!]
\centering
\begin{tabular}{c | c c} 
\hline
 Sequence Length & Train Time (minutes) & Test Time (seconds) \\ [0.5ex] 
 \hline
50 words &7:28 & $<$0.001 \\ 
100 words  & 7:40 & $<$0.001 \\
200 words & 7:57 & $<$0.001 \\
250 words & 8:03 & $<$0.001 \\
300 words & 8:18 & $<$0.001 \\
350 words & 8:30 & $<$0.001 \\
400 words & 8:39 & $<$0.001 \\ 
450 words & 8:44 & $<$0.001 \\ 
500 words & 8:49 & $<$0.001 \\
600 words & 8:55 & $<$0.001 \\
1000 words &9:03 & $<$0.001 \\
 \hline
\end{tabular}
\caption{Comparison of the sequence length of a LTM on the PTB dataset on a 1080i GPU computer}
\label{table:25}
\end{table}

LTM can be run on limited resources, with the use of only a CPU with a low RAM of 4 GB and not be affected except in the reduction on training time. Therefore, the power usage of the LTM is lower than most of the state-of-the-art large models.

\section{Conclusion}
A Long Term Memory Network, to handles long natural language sequence without being affected by the vanishing or exploding gradient problem is introduced. LTM introduces a different cell architecture. LTMs gates are used to carry on the past sequence rather than to forget the past sequence. This allows the LTM carry forward long sequences. LTM is tested on language modeling tasks and it has outperform the popular memory networks in long term dependency and have shown comparable results in short term dependencies. LTM was tested on Penn Tree bank dataset, Google Billion Words dataset and WikiText-2. Furthermore, LTM has shown that it can converge fast and does not get over trained. Each gate impact is analysed on learning short term dependencies and long term dependencies.

\ifCLASSOPTIONcaptionsoff
  \newpage
\fi



%

\bibliography{mybibfile}
\bibliographystyle{ieeetr}

%




\end{document}